\documentclass[final]{ws-ijait}

\usepackage{cite}
\usepackage{amsmath,amssymb,amsfonts}
\usepackage{algorithmic}
\usepackage{graphicx}
\usepackage{textcomp}
\usepackage{xcolor}
\usepackage{verbatim}
\usepackage{alltt}
\usepackage{url}

\usepackage{booktabs}
\usepackage{ifthen}
\usepackage{framed} 
\usepackage{stmaryrd}
\usepackage{multirow}
\usepackage{latexsym}
\usepackage{mathpartir}
\usepackage{listings}
\usepackage[caption=false, font=footnotesize]{subfig}
\usepackage{flushend}

\usepackage[ruled,vlined]{algorithm2e}

\newtheorem{Ex}{Example}
\newtheorem{Def}{Definition}

\usepackage{todonotes}

\def\ftrue{\ensuremath{\mathalpha{\textrm{\upshape{\texttt{true}}}}}}
\def\ffalse{\ensuremath{\mathalpha{\textrm{\upshape{\texttt{false}}}}}}

\def\fin{\ensuremath{\mathrel{\textrm{\upshape{\texttt{in}}}}}}

\def\fcn{\ensuremath{\mathalpha{\textrm{\upshape{\texttt{cn}}}}}}
\def\fcd{\ensuremath{\mathalpha{\textrm{\upshape{\texttt{cd}}}}}}
\def\fcxd{\ensuremath{\mathalpha{\textrm{\upshape{\texttt{cxd}}}}}}
\def\fite{\ensuremath{\mathalpha{\textrm{\upshape{\texttt{ite}}}}}}
\def\fci{\ensuremath{\mathalpha{\textrm{\upshape{\texttt{=>}}}}}}

\def\feq{\ensuremath{\mathalpha{\textrm{\upshape{\texttt{\#=}}}}}}
\def\fneq{\ensuremath{\mathalpha{\textrm{\upshape{\texttt{\#\textbackslash =}}}}}}
\def\flt{\ensuremath{\mathalpha{\textrm{\upshape{\texttt{\#<}}}}}}
\def\fleq{\ensuremath{\mathalpha{\textrm{\upshape{\texttt{\#=<}}}}}}
\def\fgt{\ensuremath{\mathalpha{\textrm{\upshape{\texttt{\#>}}}}}}
\def\fseq{\ensuremath{\mathalpha{\textrm{\upshape{\texttt{\#>=}}}}}}

\def\bnfeq{\ensuremath{\mathrel{\mathrel{::}=}}}
\def\bnfsep{\ensuremath{\mathrel{|}}}

\def\envantiresl{\mathrel{\rlap{\raise.0ex\hbox{$-$}}{\vartriangleleft}}}

\usepackage{tikz}
\usetikzlibrary{arrows}
\tikzstyle{oval}=[circle,draw,inner sep=0pt,minimum size=10mm]
\tikzstyle{rond}=[circle,draw,inner sep=0pt,minimum size=2mm]

\newcommand{\clpfd}{\texttt{clpfd}\xspace}
\newcommand{\smt}{\texttt{smt}\xspace}
\newcommand{\X}{\mathcal{X}}
\newcommand{\C}{\mathcal{C}}
\newcommand{\D}{\mathcal{D}}
\newcommand{\CSP}{(\X, \C, \D)}

\definecolor{flash}{rgb}{0.80,0,0}
\definecolor{flashb}{rgb}{0,0,0.80}

\def\BibTeX{{\rm B\kern-.05em{\sc i\kern-.025em b}\kern-.08em
    T\kern-.1667em\lower.7ex\hbox{E}\kern-.125emX}}
\begin{document}

\markboth{Gotlieb, Marijan and Spieker}
{ITE: A Lightweight Implementation of Stratified Reasoning for Constructive Logical Operators}

\title{ITE: A Lightweight Implementation of Stratified Reasoning for Constructive Logical Operators}

\author{Gotlieb, Arnaud}
\address{\textit{Simula Research Laboratory} \\
Lysaker, Norway\\
arnaud@simula.no}

\author{Marijan, Dusica}
\address{\textit{Simula Research Laboratory} \\
Lysaker, Norway\\
dusica@simula.no}

\author{Spieker, Helge}
\address{\textit{Simula Research Laboratory} \\
Lysaker, Norway\\
helge@simula.no}

\maketitle

\begin{history}
\received{(25 April 2019)}
\revised{(12 April 2020)}
\end{history}

\begin{abstract}
  Constraint Programming (CP) is a powerful declarative programming paradigm where inference and search are interleaved to find feasible and optimal solutions to various type of constraint systems. However, handling logical connectors with constructive information in CP is notoriously difficult. This paper presents \underline{I}f \underline{T}hen \underline{E}lse (ITE), a lightweight implementation of stratified constructive reasoning for logical connectives. Stratification is introduced to cope with the risk of combinatorial explosion of constructing information from nested and combined logical operators. ITE is an open-source library built on top of SICStus Prolog \clpfd, which proposes various operators, including constructive disjunction and negation, constructive implication and conditional. These operators can be used to express global constraints and to benefit from constructive reasoning for more domain pruning during constraint filtering. Even though ITE is not competitive with specialized filtering algorithms available in some global constraints implementations, its expressiveness allows users to easily define well-tuned constraints with powerful deduction capabilities. Our extended experimental results show that ITE is more efficient than available generic approaches that handle logical constraint systems over finite domains.
\end{abstract}

\keywords{Constraint Programming; Constructive Disjunction; Stratified Reasoning.}

\section{Introduction}
Constraint Programming (CP) is a powerful programming paradigm which is useful to model and solve difficult combinatorial problems \cite{HCP06}. The paradigm has proved successful to tackle various real-world problems, spanning from car manufacturing, industrial robotics, configuration design to energy production and hardware and software engineering. As compared with other paradigms such as Linear Programming or Dynamic Programming, one of the key advantages of CP is related to its expressivity. CP usually embeds a rich constraint language expressing several types of relations, including mathematical, logical, and symbolic constraints, as well as various computational domains, such as bounded subsets of integers, reals, strings, lists, graphs, etc. 

However, unlike SAT-solving or SMT-solving, CP is intrinsically designed to deal only with conjunctions of constraints \cite{HCP06}. Even though powerful deduction capabilities are implemented in CP, disjunctions are usually treated in a simple way by separating the disjuncts and solving them separately without trying to reconstruct some combined information from disjuncts. Similar approaches have been depicted for negation, implications and conditionals. So, in traditional CP, a constraint satisfaction problem (CSP) is composed of a set of constraints, implicitly combined via conjunction, and a set of variables, each of them having a finite variation domain. Dealing efficiently with general disjunction in CP has always been perceived as challenging, because of the uneasy but necessary trade-off between inference and search. On the one hand, inference tries to deduce as much information as possible from the constraints themselves, while search, by making hypothesis, explores a large search space where constraints are used to prune inconsistent branches. In this process, disjunctions just render the process inefficient as they cannot be used to prune subtrees of the search tree. Noticeable extensions include the definition of global constraints which embed specialized and efficient filtering algorithm for disjunctive reasoning, such as \texttt{element} \cite{HSD98}, \texttt{nvalue} \cite{PR99}, \texttt{cumulative} \cite{Beldiceanu2011},  \texttt{ultrametric} \cite{MP08} or \texttt{disjunctive} \cite{FQ18}. 

To illustrate the lack of inference for general disjunctive constraints, let us consider the following request\footnote{All the queries of the paper can be run using SICStus Prolog version 4.5.0, augmented with the ITE library} to SICStus Prolog \clpfd \cite{COC97}, a state-of-the-art CP solver over Finite Domains:
\begin{Ex}(\verb+#\/+ is the logical or operator of \clpfd, based on reification)\\
\verb$Query: Y in 62..77, (X#=6) #\/ (X#=13) #\/ (X#=Y).$\\
\verb$Answer: X in inf..sup$
\end{Ex}
In this example, there is no available information on the FD\_variable {\tt X}, then the solver cannot refute any of the disjuncts and it cannot perform any pruning. As a consequence, the domain of {\tt X} is left unconstrained (i.e., \verb$inf..sup$ where \verb+inf+ and  \verb+sup+ denote $-\infty$ and $+\infty$). 
For this request, a stronger expected answer would be as follows:
\begin{Ex}($\fcd$, available in ITE,  means \underline{c}onstructive \underline{d}isjunction)\\
\verb$Q: Y in 62..77, (X#=6) cd (X#=13) cd (X#=Y).$\\ %
\verb$A: X in{6}\/{13}\/(62..77)$
\end{Ex}  
In order to get this result, using some global reasoning through a {\it constructive disjunction} operator is necessary. Constructive disjunction is an operator that can construct common knowledge from both disjuncts, without knowing which one will eventually be true. In most CP implementations, there is no constructive disjunction operator natively available, as it usually is considered too costly to propagate \cite{JMN10}.

Note that, besides the simple example given above, more complex constraints can be combined using logical operators, such as in the following example:
\begin{Ex}($\fcn$ means \underline{c}onstructive \underline{n}egation)\\
  \verb$Q: A in 1..10, B in 1..10,$\\
  \verb$   (A#>1,B#<9)cd(A#>2,B#<10), (A+7#=<B)cd(cn(B+7#>A)).$\\
\verb$A: A in 8..10, B in 1..3$
\end{Ex}     
Constructive reasoning requires to propagate domain reduction over each disjunct before any constructive information can be propagated through the constraint network.  Such constructive information is often crucial to solve efficiently practical real-world problems originating from planning \cite{Do01}, scheduling \cite{MGS17}, industrial robotics \cite{MGH15}, software engineering \cite{ANB13} or configuration. However, it comes at a certain cost which needs to be controlled in order to deal with an inevitable combinatorial explosion when the number of disjunctive operators grows.

In \cite{GMS18}, we introduced some ideas on how to implement constructive disjunction on top of SICStus Prolog \clpfd\ \cite{COC97} and presented initial experimental results. In the present paper, we extend our lightweight approach to constructive operators and provide an open-source and open-access library called \underline{I}f \underline{T}hen \underline{E}lse  (ITE)\footnote{Available at  \url{https://github.com/ite4cp/ite}}.

ITE includes constraints such as constructive disjunction, negation, implication, conditional, where parametric {\it stratified reasoning} is available to cope with the inherent combinatorial explosion of disjunctive reasoning. By comparing various operators which handle disjunctions in SICStus Prolog \clpfd, we show that ITE is a good compromise between inference and search to solve problems for which specialized global constraint is not available. In particular, we propose lightweight implementations of several useful global constraints with ITE which demonstrate the versatility and expressiveness of the library as compared to existing methods.

The paper is organized as follows. Next section reviews related works while Section \ref{Sec:Back} introduces necessary notations and CSP background to understand the rest of the paper. Section \ref{Sec:Cons} details the constructive disjunction and constructive negation operators and discusses of their implementation. This section introduces various implementations of constraint entailment as well, and shows several examples to ease the understanding of relaxed versions of entailment. Section \ref{Sec:Strat} details stratified reasoning over constructive operators to cope with the combinatorial explosion of constraint propagation in the implementation of these operators. Section \ref {Sec:GlobalCtr} shows how to implement existing global constraints with constructive operators and recursion. Section \ref{Sec:Impl} presents ITE, our SICStus Prolog library constructed on top of \clpfd implementing constructive operators. This section also details experimental results obtained with ITE, while Section \ref{Sec:Conc} concludes the paper and discusses further work.

\section{Related Work}\label{Sec:Related}
{\bf Constructive disjunction is an old topic in CP}. After a pioneering proposition by Van Hentenryck \cite{HSD98}, several implementations of constructive disjunction have been proposed in different CP solvers such as Oz \cite{WM96}, Gecode \cite{SS08}, SICStus Prolog \cite{BCF13} or Choco4 \cite{choco}. The implementation of constructive disjunction is straightforward when it is considered as an extension of {\it reification} \cite{LS09}. By associating to any constraint\ $C$, a Boolean variable\ $b$ that represents the truth value of the constraint (noted $b \leftrightarrow C$), logical clauses over these Boolean variables can be used to lazily propagate the disjunction of constraints. For example, this approach is implemented in Choco4, where an embedded SAT solver can propagate logical combination of constraints. In SICStus Prolog \clpfd, besides a simple constraint reification scheme available for elementary constraints and some global constraints, a more sophisticated approach is implemented in the {\tt smt} global constraint \cite{BCF13}. The {\tt smt} global constraint propagates disjunctive combination of linear constraints. Reasoning over disjunctions of temporal constraints has also often been considered in CP, with specialized algorithms \cite{TP03}.

The problem with reification is related to its weakness to propagate strong constructive information. In fact, constraint entailment is limited by its exclusive usage of local reasoning, that is, by filtering only over the local variables of the reified constraint. In most cases, propagating more constructive content is possible but the implementation of constructive disjunction requires to find a good tradeoff between strong inference and strong search. An interesting attempt in that direction is proposed in \cite{TC07,Neveu2006} where interval reasoning propagates more information through constraint refutation. This approach is however limited to numerical constraint systems, that is, systems over real-valued variables, and does not easily extend to finite domains variables and constraints. Similarly,  \cite{Lho03} presents a novel algorithm which is as powerful as constructive disjunction, but which requires to maintain satisfying assignments for each disjunct of the constraint. This approach requires to modify the constraint solver in depth and requires in particular to modify the filtering algorithm of each propagator. Taking inspiration of SAT-solving and SMT-solving, where part of the success comes from the efficient propagation of unit clauses, \cite{JMN10} proposes to maintain the activity of at most two constraints of the disjunction at any time. By implementing new features within the solver such as satisfying sets, constraint trees and movable triggers, the approach of \cite{JMN10} successfully achieves important speedups for various disjunctive models, including the super tree problem which makes use of the ultrametric global constraint \cite{MP08}. A more lightweight method is given in \cite{BW05} where the goal is to encode various global constraints with logical connectives such as {\sc Or}, {\sc And}, {\sc Not} and characterize the cases where generalized arc-consistency can be achieved. In that respect, \cite{JMN10} generalizes the approach of \cite{BW05} by proving that their proposed filtering algorithm for {\sc Or} enforces generalized arc-consistency when all its disjunct constraints are filtered, and no variables is shared among the disjuncts. Recently, an attempt was made to propose specialized global constraints dedicated to disjunctive problems. In particular, \cite{FQ18} introduced filtering algorithms for the {\sc Disjunctive} constraint with linear-time complexity, which is a simplified version of the {\sc Cumulative} global constraint. Even though these propositions are interesting to develop a library of logical connectives, we observe that they require to modify in depth the constraint solver and/or its constraint propagators and that they do not guarantee to filter as constructive disjunction can deduce. The implementation of constructive disjunction presented in this paper has this property and can be added on top of any constraint solver. Its expressivity is also strengthened by the possible combination with recursion. Extending initial ideas presented in \cite{GMS18}, we show here that various global constraints can be interestingly implemented using stratified constructive disjunctions and recursion (see Section \ref{Sec:GlobalCtr}). The downside is however rooted in the inherent complexity of propagating constructive information which makes the approach inappropriate in a number of cases. We thus introduce stratified reasoning to deal with the inherent risk of combinatorial explosion when full constructive disjunction reasoning is used.

\section{Background}\label{Sec:Back}

\subsection{Constraint Satisfaction Problems}
A Constraint Satisfaction Problem (CSP) $\CSP$ is composed of a set of variables $\X = \{X_1,\dots,X_n\}$, a set of constraints $\C = \{C_1,\dots,C_m\}$ and a domain $\D$ which stands for the Cartesian Product $D_1 \times \dots \times D_n$, where each variable $X_i$ takes a value in the finite domain $D_i$. There are $n$ variables and $m$ constraints. A constraint $C$ is a relation over a subset of $r$ variables from $\X$ which restrains the acceptable tuples for the relation ($r$ is called the arity of $C$). The subset of variables involved in the relation, noted $var(C)$, is usually fixed in advance, but sometimes the size of the subset is initially unknown. In the former case, the relation is called a {\it primary constraint} while it is called a {\it global constraint} in the latter. $C(X_j,\dots, X_{j+r})$ is {\it satisfied}  by an assignment of its variables $X_j,\dots, X_{j+r}$ to values $v_j,\dots, v_{j+r}$ from their domain if and only if $C(v_j,\dots, v_{j+r})$ evaluates to true. A CSP $\CSP$ is {\it satisfiable} if and only if there exists at least one full assignment of its variables $\X$ such that all constraints in $\C$ are satisfied. Such an assignment is called a {\it solution}. The set of all solutions of a CSP $\CSP$ is noted $sol(\C)$ and we have, $sol(\C) \subseteq \D$. When the CSP has no solution, i.e., when there is at least one domain which is empty, then the set of constraints $\C$ is said to be {\it unsatisfiable} or {\it inconsistent}. 
\begin{Def}{\bf (Inconsistency)}\\
A CSP $\CSP$ is {\it inconsistent} if and only if $\exists i$ such that $D_i = \emptyset$ in $\D$.
\end{Def}
Solving a CSP means to either find a solution or to show inconsistency, by using local filtering consistencies. Among the many filtering consistencies which are discussed in the literature or implemented in CP solvers, we focus here only on {\em hyperarc-consistency} and {\em bound-consistency} which are mainstream.
\begin{Def}{\bf (hyperarc-consistency)}\\
For a given CSP $\CSP$, a constraint $C$ is {\it hyperarc-consistent}  if for each of its variable $X_i$  and value $v_i \in D_i$ there exist values  $v_j,\dots,v_{i-1},v_{i+1},\dots,v_{j+r}$ in $D_j,\dots,D_{i-1},D_{i+1},\dots,D_{j+r}$ such that $C(v_j,\dots,v_{j+r})$ is satisfied. A CSP $\CSP$ is hyperarc-consistent each of its constraints $C \in \C$ is hyperarc--consistent.
\end{Def}
Filtering a CSP with hyperarc-consistency, also known as generalized arc-consistency, can be demanding in terms of time or space computational resources. Thus, other consistencies, less demanding, have been proposed and, among them, bound-consistency is considered as a good compromise. 
\begin{Def}{\bf (bound-consistency)}\\
A constraint $C$ is {\it bound-consistent} if for each variable $X_i$ and value $v_i \in \{min(D_i),max(D_i)\}$ there exist values $v_j,\dots,v_{i-1},v_{i+1},\dots,v_{j+r}$ in $D^*_j,\dots,D^*_{i-1},D^*_{i+1},\dots,D^*_{j+r}$  such that $C(v_j,\dots,v_{j+r})$ is satisfied. In this definition, $D^*_j$ stands for $min(D_j) \dots max(D_j)$ (obviously $D^*_j \supseteq D_j$). A CSP $\CSP$ is {\it bound-consistent} if for each constraint $C \in \C$, $C$ is bound-consistent.
\end{Def}
In this paper, we will use a generic function name to refer to the CSP filtering process, namely a function called {\it propagate}. The function $\mbox{propagate}_{k}(C, \D) \rightarrow \D'$ takes as inputs a constraint $C$ to filter and the Cartesian product of all domains $\D$, and returns filtered domains $\D'$. This function is parametrized by $k$, which is a user-defined parameter used to define stratified reasoning. More details on CSP solving and local filtering consistencies can be found in \cite{MS98, HCP06, Lec09}  and a detailed presentation of the parameter $k$ is given in Section \ref{Sec:Strat}. Note that when $k$ is irrelevant, i.e., when there is no stratified reasoning, then $k$ is assimilated to $\infty$ and the function becomes $\mbox{propagate}_{\infty}$.

\subsection{User-Defined Global Constraint}
\label{Sec:global}
As said earlier, global constraints are relations over a non-fixed subset of variables. Among the simplest examples of such constraints, one finds {\tt all\_different$(X_1, \dots, X_n)$}  constraint \cite{Reg94}, which constrains all variables $X_1,\dots,X_n$ to take different values,  or {\tt element$(I, (X_1,\dots,X_n), V)$} which constrains $X_I$ to be equal to $V$. Modern CP solvers allows users to define is their own global constraints by using dedicated mechanisms.

Defining a global constraint in a CP solver means to give the three following elements:
\begin{enumerate}
\item {\bf (Interface)} The interface includes the constraint name,  its list of variables (possibly unbounded), and optionally, additional input parameters~;
\item {\bf (Algorithm)} A filtering algorithm is launched each time the global constraint is awaked into the propagation queue of the CP solver. This filtering algorithm aims to prune the variable domains from their inconsistent values by using the constraint semantics, the current domains of variable domains and  also,  the status of domains as they were in the previous awakening of the constraint~; 
\item {\bf (Awakening)} Deciding when to launch the filtering algorithm is crucial to control the efficiency of the solver. For that, constraint awakening conditions must be defined by saying, for example, that the constraint shall be awaked only one boundary of a variable domain has changed. Other awakening conditions can be used such as awakening the constraint when one of its variables becomes instantiated or when a specific relation is established between the variables.
\end{enumerate} 

By using a global constraint definition mechanism such as the one available in SICStus Prolog \clpfd\ \cite{COC97}, it is possible to implement various types of constraints, including those which are used to combine constraints with logical operators. In this paper, we propose a lightweight implementation of new logical connectives on top of \clpfd\, without requiring to revise deeply the architecture and implementation of the solver itself.

\subsection{Syntax of the ITE Constraint Language}\label{sec:syntax}
Fig.~\ref{fig:syntax} details the syntax of the ITE constraint language. 
\begin{figure*}
\small
\[
 \begin{array}{p{1.01cm}lll}
   $CtrBody $ & \bnfeq  & var \bnfsep \ftrue \bnfsep 1 \bnfsep \ffalse \bnfsep 0  & \textrm{}\\ 
            & \bnfsep & var\ \fin\ Range                      & \textrm{\{Domain constraint\}}\\
            & \bnfsep & Expr                                           & \textrm{\{Arithmetical Expression\}}\\
            & \bnfsep & CtrBody\ RelOp\ CtrBody      & \textrm{\{Logical constraint\}}\\
            & \bnfsep & CtrBody,\  CtrBody                  & \textrm{\{Constraint conjunction\}}\\
            & \bnfsep & \fcn (CtrBody) \bnfsep \fcn (CtrBody, Env)                    & \textrm{\{Stratified constructive negation\}}\\
            & \bnfsep & CtrBody\ \fcd\ CtrBody         & \textrm{\{Stratified constructive disjunction\}}\\
            & \bnfsep & \fcd (CtrBody, CtrBody, Env)   & \\
            & \bnfsep & CtrBody\ \fcxd\ CtrBody       & \textrm{\{Constructive exclusive disjunction\}}\\
            & \bnfsep & \fcxd (CtrBody, CtrBody, Env) & \\
            & \bnfsep & CtrBody\ \fci\ CtrBody          &  \textrm{\{Statified constructive implication\}}\\
            & \bnfsep & \fci (CtrBody, CtrBody, Env)    & \\
           & \bnfsep & \fite (CtrBody, CtrBody, CtrBody, Env)                           & \textrm{\{Statified constructive if\_then\_else\}}\\
    $RelOp$   & \bnfeq & \feq\ \bnfsep \fneq \bnfsep \flt \bnfsep \fleq \bnfsep \fgt \bnfsep \fseq\\
 \end{array}
\]
\caption{Syntax of ITE Constraint Language}\label{fig:syntax}
\end{figure*}
The ITE language introduces new operators working on existing arithmetic constraints as well as logical connectors such as $\fcn, \fcd, \fcxd, \fci, \fite$. It is worth noticing that ITE has two versions of constructive negation and constructive disjunction: the first version implements full constructive reasoning while the second implements stratified reasoning as described in Section \ref{Sec:Strat}.

\section{Constructive Operators}\label{Sec:Cons}
Implementing additional constraints such as disjunction, negation, etc. requires to reason on the truth value of the constraints and their possible entailment or disentailment \cite{CCD94}.

\begin{Def}{\bf (entailment)}\\
A constraint $C(X_j,\dots,X_{j+r})$ is {\it entailed} by a CSP $\CSP$ if and only if, for any partial assignment $v_j,\dots,v_{j+r}$ of the solutions of $\CSP$, the constraint $C(v_j,\dots,v_{j+r})$ is satisfied. On the contrary $C(X_j,\dots,X_{j+r})$ is {\it disentailed} if and only if for any partial assignment $v_j,\dots,v_{j+r}$ of the solutions of $\CSP$,  $C(v_j,\dots,v_{j+r})$ is violated.
\end{Def}
By construction, all constraints included in $\C$ are entailed by $\CSP$, but according to the definition of entailment, $C$ is not necessary part of $\C$.
\begin{Ex}\\
If all couples of $D(X_1)=\{2,3\}, D(X_2)=\{2,3\}$ are solutions of a CSP $\CSP$ then, for instance, the constraint $abs(X_1 - X_2) \leq 1$ is entailed by $\CSP$. However, the constraint $abs(X_1-X_2) > 1$ is disentailed because none of the tuples satisfy it.
\end{Ex}
Proving constraint entailment (or disentailment) is as demanding as finding all solutions of a CSP and as solving CSP is NP-hard in the general case \cite{HSD98}, relaxations of constraint entailment have to be considered. 

\subsection{Relaxations of Constraint Entailment}
Relaxations of constraint entailment in practical settings is based on local consistency filtering. In \cite{HSD98}, two partial constraint entailment tests have been proposed, namely {\it domain-entailment} and {\it interval-entailment}. These tests are based respectively on hyperarc-consistency and bound-consistency (See Section \ref{Sec:Back}), but they may not be sufficiently deductive in the case of logical connectors. Hence, we present here another partial entailment test which is based on constraint refutation, namely Abs-entailment and initially introduced in \cite{GBR00}. {\it Abs-entailment} uses the filtering capabilities of the whole CSP to try to refute the negation of the constraint to be entailed. Formally speaking, the following definitions present these three relaxations of entailment.

\begin{Def}{\bf (domain-entailment)}\\
A constraint $C(X_j,\dots,X_{j+r})$ is {\it domain-entailed} by $\CSP$ if and only if  for all domains $D_i$ in $\D = D_1 \times \dots \times D_n$  and all values $v \in D_i$, $C(v_j,\dots,v_{j+i-1},v,v_{j+i+1},\dots,v_{j+r})$ is satisfied.
\end{Def}
Using the definition, one can see that a constraint $C$ can be entailed,  but not necessary domain-entailed by a CSP $\CSP$. On the contrary, any constraint $C$ which is domain-entailed is also automatically entailed by $\CSP$. Note also that domain-entailment is very demanding as it requires to consider all values of all domains in $\D$.
\begin{Ex}\\
If $X_1 \in \{2, 3, 4\}$, $X_2 \in \{2, 3, 4\}$ and $\C=\{ X_1 \leq X_2\}$  then the constraint $X_1 \leq X_2+1$ is entailed by $\CSP$  but not domain-entailed. $X_1 \leq X_2+1$ is entailed because all satisfying tuples of $X_1 \leq X_2$, namely $\{(2,2), (2,3), (2,4), (3,3), (3,4), (4,4)\}$ are also solutions of $X_1 \leq X_2+1$. But, the constraint is not domain-entailed because there are pairs in $(\{2,3,4\},\{2,3,4\})$ which do not satisfy $X_1 \leq X_2+1$. Consider for example the pair $(4,2)$.
\end{Ex}
Domain-entailment requires only to examine the satisfying tuples of $C$ with respect to the current domains of its variables and as such it is a local property. However, it requires to examine all tuples in the domains which can be very demanding when the domains are composed of many values. Another relaxation has been proposed to cope with that issue, namely {\it interval-entailment}.
\begin{Def}{\bf (interval-entailment)}\\
A constraint $C(X_j,\dots,X_{j+r})$ is {\it interval-entailed} by a CSP $\CSP$ if and only if for all domains $D_i$ and for all values $v$ in $min(D_i)\dots max(D_i)$, the constraint $C(v_j,\dots,v_{j+i-1},v,v_{j+i+1},\dots,v_{j+r})$ is satisfied.
\end{Def}
Of course, if $C$ is interval-entailed by a CSP  then $C$ is also domain-entailed as $D_i \subseteq min(D_i)..max(D_i)$ but the opposite if false. Nevertheless, these two relaxations of entailment are quite similar and both have been used in CP solvers to implement classical (non-constructive) disjunction and negation. In SICStus Prolog \clpfd\, both relaxations are used for implementing reification operators of various constraints. As said earlier, reification is used to evaluate the truth value of constraint and propagating this value in logical combinations of constraints, including within some global constraints \cite{BCF13}.  
A third partial entailment test, based on constraint refutation, is presented here~:
\begin{Def}{\bf (abs-entailment)}\\
A constraint $C$ is {\it abs-entailed} by a CSP $\CSP$ if filtering the modified CSP $(\X, \C \wedge \neg C, \D)$ yields to an inconsistency. Formally, $C$ is {\it abs-entailed} if and only if $\mbox{propagate}_{\infty}(\neg C, \D) = \emptyset$
\end{Def}
The definition is operational as it is based on concrete filtering procedures available in most CP solvers. Provided that {\it hyperarc-consistency}  and {\it bound-consistency} are sound relaxations of consistency, we get that {\it abs-entailment} is a sound relaxation of entailment if it uses these local filtering properties. Indeed, if $\C \wedge \neg C$ is inconsistent then it means that $\C$ entails $C$. However, unlike {\it domain-entailment} or {\it interval-entailment}, checking {\it abs-entailment} is not restricted to evaluating the satisfying tuples of $C$. It requires to filter the whole CSP $\CSP$ augmented with $\neg C$. Moreover, it requires to backtrack to the initial computational state after having check the inconsistency of $(\X, \C \wedge \neg C, \D)$. Thus, in some cases, it may require to call recursively the solver which is inappropriate for some CP solvers. Nevertheless, {\it abs-entailment} is a powerful entailment check and much more deductive than {\it domain-entailment} and {\it interval-entailment}. Our implementation of constructive disjunction and negation in ITE are based on this partial entailment check.

\subsection{Constructive Disjunction}
By using a global constraint definition mechanism, we can introduce a new constructive disjunctive operator, namely $C_1\ \fcd\ C_2 $, with Algorithm \ref{alg:cd}.
\begin{algorithm}
$(\D^{r}: Dom, S: Status) \leftarrow \fcd(C_1: Constr, C_2: Constr, \D: Dom)$ \\
$\D_1 \leftarrow \mbox{propagate}_{\infty}(C_1, \D)$\\  
$\D_2 \leftarrow \mbox{propagate}_{\infty}(C_2, \D)$\\  
\Case{$\D_1=\emptyset$ and $\D_2=\emptyset$} {\Return{($\emptyset$, fail)}}
\Case{$\D_1=\emptyset$} {\Return{($\D_2$, exit)}} 
\Case{$\D_2=\emptyset$} {\Return{($\D_1$, exit)}}
\Else{\Return{($\D_1 \cup \D_2$, suspend)}}
\caption{A Generic Constructive Disjunction Algorithm}
\label{alg:cd}
\end{algorithm}
Algorithm \ref{alg:cd} makes use of the propagate function which is provided for filtering any constraint system. It is obviously parametrized by the level of consistency offered by the solver and its implemented constraints. Note that testing if $D_1$ or $D_2$ is void corresponds actually to an implementation of {\it abs-entailment} for the constructive disjunction. Indeed, testing if $D_1 = \emptyset$ (resp. $D_2$) is like checking if $\neg C_1$ is abs-entailed (resp. $C_2$).  Algorithm \ref{alg:cd}  implements a monotonic propagator as it returns filtered domains $\D^{r}$ which are necessarily a subset of $\D$, i.e., $\D^{r} \subseteq \D$. One can see that only $D_1$, or $D_2$, or else $D_1 \cup D_2$ can be returned. In addition to the filtered domains, Algorithm \ref{alg:cd} also returns a status of the $\fcd$ constraint, which can be either  $suspend, exit$ or $fail$, as explained in Section \ref{Sec:global}.

The algorithm works over $\D$  as a whole here but it can work over any projection of $\D$ on a subset of its domains. In particular, a restricted version of this algorithm can project $\D$ over the union of variable domains of $C_1$ and $C_2$, namely $var(C_1 \cup C_2)$. This allows us to preserve the local property of constraint filtering. Note that  the two successive calls to $propagate$ show that the solver is actually called twice for each $\fcd$ operator, which can lead to a combinatorial explosion of the number of calls to the solver when the number of disjunctive constraints grows.  This raises the question of what to do when, during the recursive calls to the solver, other $\fcd$  operators are encountered. The following example illustrates the normal behavior of the generic algorithm.
\begin{Ex}\\
  \verb+Q: A in 1..5, B in 1..5, C in 1..5,+\\
  \verb+Q  (A-B#=4)cd(B-A#=4),(A-C#=4)cd(C-A#=4).+\\
\verb+A:  A,B,C in {1}\/{5}+\\
\end{Ex}
The values {\tt 2,3,4} are pruned from the domains of {\tt A,B,C} because any pairwise combination of the constraints will prune these values, but this powerful deduction has been obtained by solving {\tt 4} constraint systems in this example. This nested analysis may lead to very precise results but, at the price of being computationally expensive. In Section \ref{Sec:Strat}, we introduce stratified reasoning as a way to cope with the possible combinatorial explosion of this reasoning, but let us examine first the other logical operators using constructive reasoning.
 
\subsection{Constructive Negation}
In Logic Programming, usual negation operators implement {\it negation-as-failure}, which triggers failure when the negated goal succeeds and conversely. Although useful in many contexts, this {\it negation-as-failure} operator coincides with logical negation only when the negated goal is grounded, which means that all its variables are bounded. In the general case, when negating a constraint, there are several unbounded variables and then posting the negation of that constraint cannot be handled anymore with the {\it negation-as-failure} operator. Constructive negation in Constraint Logic Programming has been proposed to cope with this issue, using the Clarke's completion process \cite{Fag97}. Unfortunately, for CSPs, only a few implementations have been considered. In ITE, we propose an implementation called $\fcn$, which is based on a combination of negation-as-failure, De Morgan's rewriting rules and constructive disjunction.

The following rewriting rules are considered:\\
$\fcn(true) \longrightarrow \ffalse;\ \ \ \fcn(\ffalse) \longrightarrow \ftrue$\\
$\fcn(C) \longrightarrow \ffalse \mbox{ when } var(C) = \emptyset \mbox{ and } C \mbox{ is } \ftrue$\\
$\fcn(C) \longrightarrow \ftrue \mbox{ when } var(C) = \emptyset \mbox{ and } C \mbox{ is }\ffalse$\\
$\fcn(var\ \fin\ R) \longrightarrow var\ \fin NegR$ where $NegR$ is the complementary of $R$\\
$\fcn(E_1\ RelOp\ E_2) \longrightarrow E_1\ NegOp\ E_2$  where $NegOp$ is the negation of $RelOp$\\
$\fcn((C_1, C_2)) \longrightarrow \fcd(\fcn(C_1),\fcn(C_2))$\\
$\fcn(\fcd(C_1,C_2)) \longrightarrow (C_1, C_2)$\\
$\fcn(\fcxd(C_1, C_2)) \longrightarrow \fcd((C_1, C_2), (\fcn(C_1), \fcn(C_2))$\\
$\fcn(C_1 \fci C_2) \longrightarrow (C_1, \fcn(C_2))$\\
$\fcn(ite(C, C_1, C_2, _)) \longrightarrow (\fcd(\fcn(C),\fcn(C_1)), \fcd(C,\fcn(C_2)))$\\
This approach is useful for unary operators such as constructive negation, as it allows us to eliminate the negation from any request. For the language given in Fig.\ref{fig:syntax}, it is not difficult to see that by pushing negation on elementary constraints, only negated elementary constraints remain and that they can easily be eliminated by negating the corresponding operators. For a more extended language including added global constraints, this approach would require to include the negation of global constraints which is more problematic, even though approaches exist \cite{LAG12,BCF13}.

\subsection{Other Operators}
Other logical connectors, which are  on based on constructive disjunction, can be implemented by using similar principles. In particular, exclusive disjunction (noted $\fcxd$), logical implication ($\fci$) and conditional ($\fite$) are easy to derive from the implementation of constructive disjunction.  For exclusive disjunction $C_1 \fcxd\ C_2$,  when performing the abs-entailment test for each disjunct, e.g., $C_1$, we can propagate the negation of this disjunct in addition to the other constraint, i.e., we can add $\mbox{propagate}((\fcn(C_1) ,C_2),\ \D)$. Adapting Algorithm \ref{alg:cd} for $\fcxd$ is thus not too difficult. Logical implication can be translated as follows: $C_1 \fci C_2$ is equivalent to $\fcn(C_1) \fcd\ C_2$, but such a transformation replaces one binary operator ($\fci$) by two operators ($\fcn, \fcd$). Even though this approach is certainly acceptable, in CP solving, it is interesting to keep the number of operators as low as possible. That is why we preferred, in this case, an approach where $\fci$ is implemented as a new operator with its own filtering algorithm based on a variation of Algorithm \ref{alg:cd}. Note that, as CSP works on finite domains,  the {\it closed world hypothesis} guarantees that there is no solution available which satisfies both $C$ and $\fcn(C)$. Similarly any conditional operator such as $\fite(C, C_1, C_2)$, representing  if($C$) then $C_1$ else $C_2$, can trivially be converted into an exclusive disjunction operator by using the equivalent formulae: $(C, C_1) \fcxd\ (\fcn(C), C_2)$. But, here again, we preferred a dedicated implementation of its filtering algorithm. The following example illustrates the usage of $\fite\ $ when it is combined with other constraints.
\begin{Ex}\\
\verb+Q:ite(I0#=<16, J2#=J0*I0, J2#=J0, ENV), J2#>8, J0#=2.+\\
\verb+A: J0 = 2, I0 in 5..16, J2 in 10..32+
\end{Ex}

\section{Stratified Reasoning}\label{Sec:Strat}
In the previous section, we presented implementations of logical operators without paying attention to the computational cost of propagating disjunctive constraints. Even though filtering by hyperarc-consistency and bound-consistency is time-polynomial for elementary constraints in  the worst case, the number of inconsistency checks performed can rapidly explode and lead to some unwanted combinatorial explosion. Indeed, the number of abs-entailment tests grows exponentially with the number of $\fcd$ in the formulae. To cope with this problem, we propose {\it stratified reasoning} which aims at exploring disjuncts up to a certain depth, controlled by the user. This process is stratified as it allows the user to increase the depth of the exploration.

In concrete implementation terms, by setting up a user-defined parameter $k$ to a positive integer (usually a small value), one can decrease $k$ by one each time an inconsistency check (abs-entailment test) is performed.  When $k$ reaches the value $0$, then the abs-entailment test is simply discarded, avoiding so to perform an uncontrolled and redundant exploration of all disjunctions. A straightforward algorithm based on this idea is given in Algorithm \ref{alg:scd}. In this algorithm, the variable $Env$ captures general information about the computational environment, including the value of $k$. While performing the abs-entailment test with the $propagate$ function, $k$ is simply decreased. 

This way of handling disjunctive reasoning is correct and does not compromise the final result of CP solving: it just delays full reasoning up to the final labeling stage. In fact, when there are no variables left in a disjunct, i.e., when all the variables have been labelled in one disjunct, then one can discard the $\fcd$ operation by evaluating that disjunct. This explains the presence of tests $vars(C_i) \neq \emptyset$ in Algorithm \ref{alg:scd}.
\begin{algorithm}
  $(\D^{r}: Dom, S: Status) \leftarrow \fcd(C_1, C_2, \D: Dom, Env: Environment)$ \\
  $k \leftarrow get\_k(Env)$\\
  \If{$k=0$ and ($vars(C_1) \neq \emptyset$ and $vars(C_2) \neq \emptyset$)}{\Return{($\D$, suspend)}}
  \Else{
$\D_1 \leftarrow \mbox{propagate}_{k-1}(C_1, \D)$\\  
$\D_2 \leftarrow \mbox{propagate}_{k-1}(C_2, \D, k-1)$\\
\Case{$\D_1=\emptyset$ and $\D_2=\emptyset$} {\Return{($\emptyset$, fail)}}
\Case{$\D_1=\emptyset$} {\Return{($\D_2$, exit)}} 
\Case{$\D_2=\emptyset$} {\Return{($\D_1$, exit)}}
\Else{\Return{($\D_1 \cup \D_2$, suspend)}}
  }
\caption{Constructive Disjunction with Stratified Reasoning}
\label{alg:scd}
\end{algorithm}
The following examples illustrates how stratified reasoning can benefit to the deduction capabilities of the solver. By setting up different values of $k$ (parameter selected by the user), one gets distinct results with the same constraint system. Predicates \verb+init_env/2+ and \verb+end_env/1+ are used to set up the environment for solving the formulae with ITE. Note that, in this example, when the parameter is equal to $3$, one gets the maximal possible deduction as there is no possible further deduction. So, the value $k=3$ is maximal for that formulae.
\begin{Ex}\\
  \verb+Q:init_env(E,[kflag(3)]),cd(cd(X=0,cd(Y=4,Y=5,E),E),X=9,E),+\\
  \verb+  cd(cd(Y=9,Y=6,E),cd(Y=2,Y=7,E), E), end_env(E).+\\
\verb+A:  X in {0}\/{9}, Y in {2}\/(6..7)\/{9}+\\
\\
\verb+Q:init_env(E,[kflag(2)]),cd(cd(X=0,cd(Y=4,Y=5,E),E),X=9,E),+\\
\verb+  cd(cd(Y=9,Y=6,E),cd(Y=2,Y=7,E), E), end_env(E).+\\
\verb+A:  X in inf..sup, Y in {2}\/(6..7)\/{9}+\\
\\
\verb+Q:init_env(E,[kflag(1)]),cd(cd(X=0,cd(Y=4,Y=5,E),E),X=9,E),+\\
\verb+  cd(cd(Y=9,Y=6,E),cd(Y=2,Y=7,E), E), end_env(E).+\\
\verb+A:  X in inf..sup, Y in inf..sup+
\end{Ex}

Formalizing the concept of maximal $k$, we have the following definition:
\begin{Def}{\bf (maximal k)}\\
For a given CSP $\CSP$ and a constraint $C$ of $\C$, the parameter $k$ is {\it maximal}  if and only if $k$ is the greatest value such that $\mbox{propagate}_{k+1}(C, \D)=\mbox{propagate}_{k}(C, \D)$ and $\mbox{propagate}_{k}(C, \D) \neq \mbox{propagate}_{k-1}(C, \D)$.  
\end{Def}
Note that $k$ is unrelated to the level of consistency achieved and that finding maximal $k$ does not guarantee to achieve any kind of consistency. In the example, when $k=2$ the result is neither domain- nor bound- consistent. We conjecture that maximal $k$ can be approached by using the immediate successor of the maximum depth of any disjunctive constraint in the formulae but there is no available proof. We discuss the conjecture in Section \ref{Sec:Impl} of the paper. Stratified reasoning appears to be a good trade-off between search and inference, although coupled with a decision of which aspect is more important.

\section{Global Constraints with ITE}\label{Sec:GlobalCtr}
The ITE library can be used to implement various global constraints. In this section, we show a selection of various global constraints with their ITE implementation. The objective is to show the expressiveness of the constructive operators, in particular when they are coupled with recursion.
\subsection{{\sc Ultrametric} constraint}
The {\sc Ultrametric} constraint, as proposed in \cite{MP08}, is a ternary constraint defined as follows: $$X>Y=Z \lor Y>X=Z \lor Z>X=Y \lor X=Y=Z$$. Its implementation in ITE is straightforward:
\begin{Ex}{\bf ({\sc Ultrametric} constraint)}\\
\verb+um3(X, Y, Z, ENV):-+\\
\verb+  cd(cd((X#>Y,Y=Z),(Y#>Z,X=Z),ENV),cd((Z#>X,X=Y),(X=Y,Y=Z),ENV),ENV).+
\end{Ex}
Note that balancing the $\fcd$ connectors is advantageous for ITE so that the depth of the syntax tree of formulae remains minimal.

\subsection{{\sc Domain} Constraint}\label{sec:domain}
The {\sc Domain} constraint is defined over a set of $0/1$ variables as follows: {\sc Domain}$(X,[X_1, \dots, X_n])$ is true when  $\forall i, (X = i \Leftrightarrow X_i = 1)$. Note that the {\sc Domain} constraint is natively available in SICStus Prolog \clpfd\ with {\tt bool\_channel(L, X, \#=, 0)}. Its implementation in ITE shows the combined usage of constructive disjunction and recursion:
\begin{Ex}{\bf ({\sc Domain} constraint)}\\
\verb+domctr(X, L, ENV) :-+\\
\verb+  length(L, N),  X in 1..N, domain(L, 0,1),+\\
\verb+  domctr(N, X, L, ENV).+\\
\\
\verb+domctr(1, X, [X1], ENV) :- !, X=1, X1=1.+\\
\verb+domctr(N, X, [X1|L], ENV) :- N>2, !,+\\
\verb+  N1 is N-1,+\\
\verb+  cd((X=1,X1=1,domain(L,0,0)),+\\
\verb+     (X#>1,X1=0,Y in 1..N1,incr(X,Y), domctr(N1,Y,L,ENV)),ENV).+\\
\\
\verb|incr(X, Y)  +: |\\
\verb|  X in dom(Y)+1,|\\
\verb|  Y in dom(X)-1.|
\end{Ex}
Using {\tt domctr} in one of the disjuncts of the {\tt cd} constraint illustrates the interest of ITE in terms of expressiveness. In fact, constraint reasoning can easily be combined with recursion to create powerful implementations of global constraints. Even though they have less filtering capabilities than specialized algorithms, these implementations can help the developer to create very quickly a dedicated constraint which has interesting deduction capabilities. Note that we created our own constraint $X=Y+1$ with {\tt incr(X,Y)}  in order to reach domain-consistency.

\subsection{{\sc Element} Constraint}\label{sec:element}
We give here an implementation of the well-known {\sc Element} constraint \cite{HSD98}, which is based on very similar principles as {\sc Domain}:
\begin{Ex}{\bf ({\sc Element} constraint)}\\
\verb+elemctr(I, L, J, ENV):-+\\
\verb+  length(L, N),+\\
\verb+  I in 1..N,+\\
\verb+  elemctr(N, I, L, J, ENV).+\\
\\
\verb+elemctr(1, I, [X1], J, _ENV) :- !, I=1, J#=X1.+\\
\verb+elemctr(N, I, [X1|L], J, ENV) :- N>2, !,+\\
\verb+  N1 is N-1,+\\
\verb+  cd((I=1,X1#=J),(I#>1, I1 in 1..N1, incr(I,I1),+\\
\verb+                        elemctr(N1,I1,L,J,ENV)),ENV).+ 
\end{Ex}

\subsection{{\sc Lex} Constraint}\label{sec:lexctr}
The {\sc Lex}$([X_1,\dots,X_n], [Y_1,\dots,Y_n])$ constraint is true iff $X_1 < Y_1 \lor (X_1=Y_1 \land X_2 < Y_2)  \lor  (X_1=Y_1 \land X_2=Y_2 \land X_3 < Y_3),\dots$. It can be implemented in ITE as follows:
\begin{Ex}{\bf ({\sc Lex} constraint)}\\
\verb+lexctr(X, Y, ENV):-+\\
\verb+  X=[X1|Xs], Y=[Y1|Ys],+\\
\verb+  lexctr(Xs, Ys, [X1], [Y1], fail, T, ENV),+\\
\verb+  call(cd(X1#<Y1, T, ENV)).+\\
\\
\verb+lexctr([], [], _LX, _LY, TIN, TIN, _ENV) :- !.+\\
\verb+lexctr([X|Xs],[Y|Ys],LX,LY,TIN,TOUT,ENV) :-+\\
\verb+  gen_eq(LX, LY, T1),+\\
\verb+  T = cd(TIN, (X#<Y, T1), ENV),+\\
\verb+  lexctr(Xs,Ys,[X|LX],[Y|LY],T,TOUT,ENV).+\\
\\
\verb+gen_eq([X], [Y], (X#=Y)) :- !.+\\
\verb+gen_eq([X|Xs], [Y|Ys], (X#=Y, T)) :-+\\
\verb+  gen_eq(Xs, Ys, T).+
\end{Ex}
This constraint is natively available in SICStus Prolog \clpfd under the name {\tt lex\_chain}.

\subsection{{\sc Mulctr} Constraint}\label{sec:mulctr}
Unlike the previous ones, the {\sc Mulctr} is not available in the \clpfd library and can be implemented easily. The constraint {\sc Mulctr}$(N, X, Min, Max)$ is a specialized constraint which is true if and only if $X \geq Min \land (X = N \lor X = 2N \lor \dots \lor X = M.N) \land X \leq Max$.  Its implementation in ITE is as follows:
\begin{Ex}{\bf ({\sc Mulctr} constraint)}\\
\verb+mulctr(OP, N, X, Min, Max, ENV):-+\\
 \verb+ X#>=Min, X#=<Max, fd_max(X,Xmax),fd_max(N,Nmax),+\\
 \verb+ Mmax is (Xmax div Nmax),+\\
 \verb+ gen_mulctr(1, Mmax, N, X, [], L),+\\
 \verb+ mulctr1(L, T, ENV),call(T).+\\
\\
\verb+gen_mulctr(M, M, _N, _X, L, L):-!.+\\
\verb+gen_mulctr(M, Mmax, N, X, T, [X#=M*N|Xs]):-+\\
\verb|  M1 is M+1,|\\
\verb+  gen_mulctr(M1, Mmax, N, X, T, Xs).+\\
\\
\verb+mulctr1([CTR], CTR, _ENV) :-!.+\\
\verb+mulctr1([CTR|S], cd(CTR,CS,ENV), ENV):-!,+\\
\verb+  mulctr1(S, CS, ENV).+\\
\end{Ex}

\subsection{{\sc Disjunctive} Constraint}\label{sec:disjunctive}
The {\sc Disjunctive}$([S_1,\dots,S_n], H)$ constraint is true iff  $(S_i + p_i \leq S_j)   \lor   (S_j + p_j  \leq S_i)$  for all pairs of task $S_i, S_j$ such that $i \neq j$, where  $p_i$ is the duration of task $S_i$ and $H$ is the horizon, i.e., $H=\sum_i S_i+p_i$. Its implementation in ITE is as follows:
\begin{Ex}{\bf ({\sc Disjunctive} Constraint)}\\
\verb+disjctr(S, P, H, ENV) :-+\\
\verb+  joinsp(S, P, SP),gen_disj(SP, G, ENV),+\\
\verb+  append(S, P, VSP), sum(VSP, #=, H),call(G).+\\
\\
\verb+joinsp([], [], []).+\\
\verb+joinsp([X|Xs], [Y|Ys], [(X,Y)|S]) :-+\\
\verb+  joinsp(Xs, Ys, S).+\\
\\
\verb+gen_disj([_], true, _ENV):- !.+\\
\verb+gen_disj([X|XL], (G1, G), ENV) :-+\\
\verb+  gen_disj1(X, XL, G1, ENV),+\\
\verb+  gen_disj(XL, G, ENV).+\\
\\
\verb+gen_disj1((Si,Pi),[(Sj,Pj)], RES,  ENV ) :-!,+\\
\verb|  RES = cd( Si+Pi#=<Sj,  Sj+Pj#=<Si, ENV). |\\
\verb+gen_disj1((Si,Pi),[(Sj,Pj)|YL], RES,  ENV):-+\\
\verb|  RES = (cd( Si+Pi#=<Sj,  Sj+Pj#=<Si, ENV),  G),|\\
\verb+  gen_disj1((Si,Pi), YL, G, ENV).+
 \end{Ex}

\section{Experimental Results}\label{Sec:Impl}
To confirm the effectiveness of the lightweight approach to constructive
disjunction, we compare the implementation of global constraints using ITE to
the default implementations from SICStus Prolog.
As benchmark scenarios, we use constraint formulations which heavily rely on
disjunctive clauses. Specifically, we use the global constraints {\sc Domain}, {\sc
  Element}, {\sc Lex} and {\sc Mulctr} as described in Section~\ref{Sec:GlobalCtr}.

\subsection{Setup}
For all constraints, we compare the implementation using
ITE's cd and cd(K) operators with an implementation using the regular operators
from \clpfd.
We also compare to the global constraints implementation of SICStus Prolog,
except for {\sc mulctr}, which is not implemented in \clpfd.

The benchmark dataset and its execution scripts are available as part of the ITE
software package\footnote{Available at \url{https://github.com/ite4cp/ite}}. All
experiments have been performed with SICStus Prolog version 4.5.1 on Linux
running on an Intel i7-6820HQ (2.70 GHz) with 16\,GB RAM.

\subsection{Experiments}

In the following, we will describe the actual experiments performed and their
results. In summary, before showing the detailed results, constraints
implemented on ITE are mostly competitive and outperforming the corresponding
implementations using the regular \clpfd operators. An implementation of the
constraints using \smt is usually not efficient, due to the limitation of \smt
to linear constraints. If available, a global constraint implementation from the
\clpfd library is most efficient. For all experiments, we combine the global
constraint with a deterministic labeling predicate to ensure the exact same
search process is performed by each implementation.

However, we do not claim that ITE can outperform a highly optimized
implementation of a specific global constraint. Instead, and this is one of the
major contributions of the ITE framework, it is a versatile and flexible
approach for constructive disjunction, which is a building block of many complex
constraint constructions, such as the global constraints which we consider in
the experiments. Therefore, the goal of the experiments, which is confirmed by
the results, is that the advantage of ITE's expressiveness does not lead to a
performance decrease, but is competitive or in some cases exceeding the
performance of regular operators.

\subsubsection{{\sc Domain}}

As the first global constraint, we employ {\sc Domain} as
described in Section~\ref{sec:domain} and its implementations in ITE, regular \clpfd,
\smt and the \clpfd global constraint.

\begin{figure}[t]
  \centering
  \input{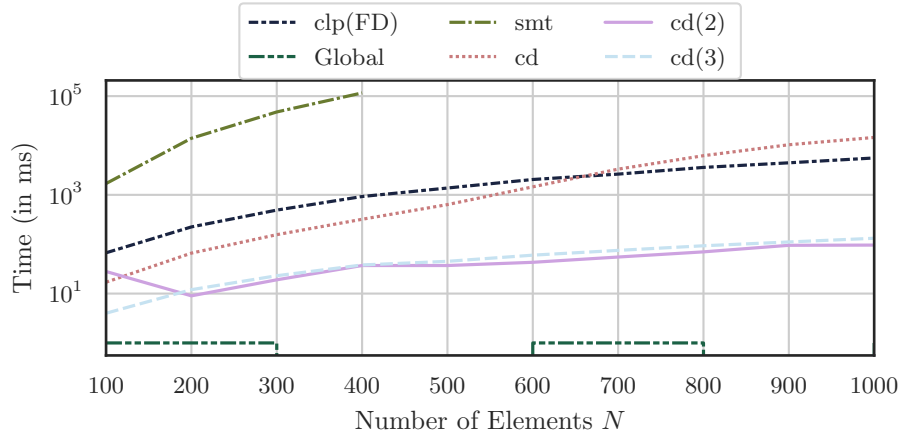}
  \caption{{\sc Domain}: Resolution time vs. number of items. y-axis: Logarithmic scale. For $N>400$ \smt exceeds the timeout of 5 minutes.}
  \label{fig:domctr}
\end{figure}

The experiment is performed by labeling an list $L$ of $N$ 0/1 variables with side constraint $X*X
< X$, such that $X$ is maximized: $length(L, N), domctr(X, L), X*X \#< N,
labeling([maximize(X)], L).$.
Labeling is used to ensure all implementations perform the same search and we
actually measure the difference in propagation performance.
We vary the number of elements $N$ from $100-1000$ with a timeout of 5 minutes.

The results are shown in Figure~\ref{fig:domctr}.
Best performance is shown by the \clpfd global constraint, as expected. This
implementation is written in native C and highly optimized, where as the other
results are based on a composition of standard constraints, either from \clpfd
or ITE.
The slow performance of \smt is explained as it does not natively support
non-linear constraints and requires an alternative formulation of the constraint.
Using ITE without an additional parameter $k$, i.e. \texttt{cd}, is competitive
to the performance of the implementation with regular \clpfd operators and
performs better when the number of elements $N$ is small to medium. 
When using an additional parameter $k$, i.e. \texttt{cd(2)} and \texttt{cd(3)},
ITE is faster to resolve the expressions than the other methods, except the
\clpfd global constraint.

\subsubsection{{\sc Element}}

\begin{figure}[t]
  \centering
  \input{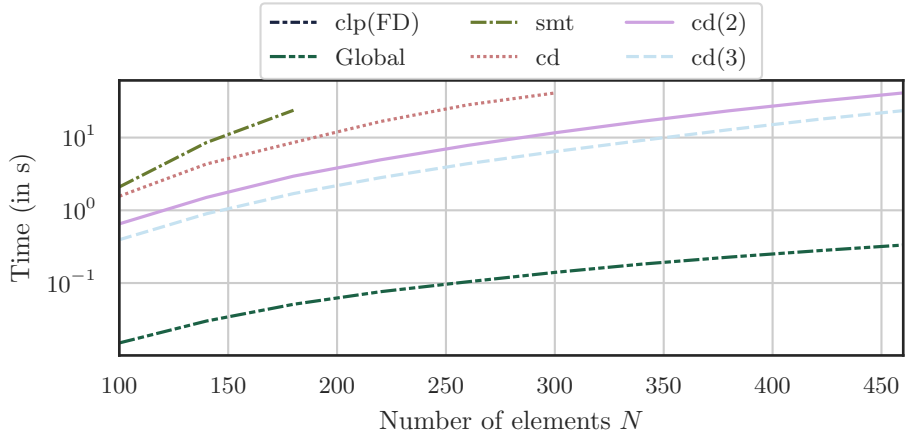}
  \caption{{\sc Element}: Resolution time vs. number of elements. y-axis:
    Logarithmic scale. \clpfd exceeds the timeout for all $N$, \smt for $N>180$
    and cd for $N>300$.}
  \label{fig:elemctr}
\end{figure}

The second global constraint is {\sc Element} as described in
Section~\ref{sec:element}, with results shown in Figure~\ref{fig:elemctr}.
A larger number of elements $N$ makes the problem more complex, such that \smt
exceeds the timeout of 1 minute for $N>180$ and \texttt{cd} for $N>300$.
\texttt{cd} with an additional parameter $k$, i.e. \texttt{cd(3)} and
\texttt{cd(4)}, are able to resolve all expressions within the time limit.
The implementation using regular \clpfd operators does not resolve any of the
expressions within the timeout, whereas the global constraint implementation of
\clpfd is the most efficient.

\subsubsection{{\sc Lex}}

The {\sc Lex} global constraint (Section~\ref{sec:lexctr}) shows interesting behaviour with a small number
of elements $N$.
With $N<30$ elements, the regular \clpfd implementation exceeds the timeout and
does not resolve the expression, but with larger $N$, this problem does not
occur and the performance is actually better than ITE's.

\begin{figure}[t]
  \centering
  \input{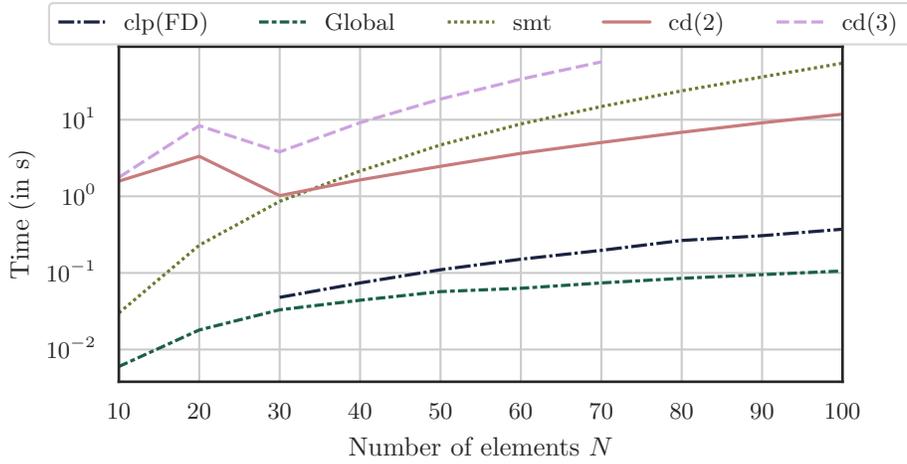}
  \caption{{\sc Lex}: Resolution time vs. number of elements. y-axis: Logarithmic scale.}
  \label{fig:lexctr}
\end{figure}

\subsubsection{{\sc Mulctr}}

Finally, for evaluation of the {\sc Mulctr} global constraint
(Section~\ref{sec:mulctr}), a larger number of $k$ values is considered ($k\in
[2;7]$). However, there is no standard implementation in \clpfd and we also do
not provide an implementation in \smt. Therefore, this experiment mostly focuses
on the differences within ITE parametrization.

\begin{figure}[t]
  \centering
  \input{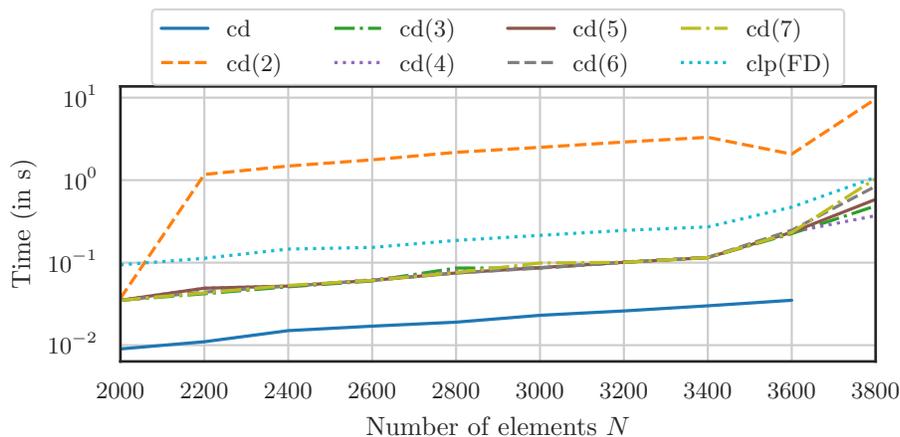}
  \caption{{\sc Mulctr}: Resolution time vs. number of elements. y-axis: Logarithmic scale.}
  \label{fig:mulctr}
\end{figure}

The results are shown in Figure~\ref{fig:mulctr} and show three main aspects:
a) a misparametrization can have a performance impact, as shown by \texttt{cd(2)};
b) once a minimum parameter value has been found, the actual performance
differences vary not much more, as shown by \texttt{cd(3)}-\texttt{cd(7)};
c) the non-parametrized \texttt{cd} is a good starting point, as it can have
good or better performance than the parametrized versions.

\section{Conclusion}\label{Sec:Conc}
In this paper, we have introduced ITE, an open-source and open-access lightweight implementation of constructive operators in the SICStus Prolog \clpfd\ solver. Our implementation provides a variety of operators which allows one to construct logical combination of constraints, including some global constraints. By exploiting abs-entailment, a relaxation of entailment, and global constraint definition mechanisms, ITE has interesting expressiveness capabilities and shows a good tradeoff between inference and search. To cope with the  risk of combinatorial explosion of constructive reasoning, our approach  provides stratified deduction, which is a versatile technique requiring only to set up a single parameter $k$. In our experiments, we have shown that ITE with stratified reasoning is an advantageous alternative to implement some global constraints. In particular, ITE is competitive with available tools that deal with disjunctive reasoning such as reification with domain- and interval- entailment or the {\tt smt} global constraint. It is however not competitive with dedicated filtering algorithms implemented in specialized global constraints. As future work, we plan 1) to propose to automatically adjust the value of $k$ for stratified reasoning with heuristics on the depth of the syntax tree of the formulae and 2) to explore the capability to infer more symbolic information from disjunctions than just pure domain information. The later proposition would require to propagate not only domains, but also deduced variable relations such as variable equality for instance. This is a challenging problem as performing the union of symbolic information in CP solving always requires to trade between precision and performance.

\section*{Acknowledgements}
This work is supported by the Research Council of Norway under the T-Largo grant agreement.
\bibliography{ijait2019}
\bibliographystyle{ws-ijait}
\end{document}